\documentclass[10pt,twocolumn,letterpaper]{article}

\usepackage[pagenumbers]{cvpr} 

\definecolor{cvprblue}{rgb}{0.21,0.49,0.74}
\usepackage[pagebackref,breaklinks,colorlinks,allcolors=cvprblue]{hyperref}
\usepackage{hyperref}
\hypersetup{colorlinks=true,urlcolor=magenta}
\usepackage{multirow}
\usepackage{diagbox}
\usepackage{graphicx}
\usepackage{makebox}
\usepackage{pifont}
\def\paperID{15029} 
\def\confName{CVPR}
\def\confYear{2025}

\title{High Dynamic Range Video Compression: A Large-Scale Benchmark Dataset and A Learned Bit-depth Scalable Compression Algorithm} 

\author{
Zhaoyi Tian, Feifeng Wang\thanks{Equal contribution}, Shiwei Wang, Zihao Zhou, Yao Zhu, Liquan Shen\thanks{Corresponding author}\\
Shangha University\\
China\\
{\tt\small \{kinda,wff0520,ieemia,yi\_yuan,yaozhu,jsslq\}@shu.edu.cn}
}

\begin{document}
\maketitle
\begin{abstract}
Recently, learned video compression (LVC) is undergoing a period of rapid development. However, due to absence of large and high-quality high dynamic range (HDR) video training data, LVC on HDR video is still unexplored. In this paper, we are the first to collect a large-scale HDR video benchmark dataset, named HDRVD2K, featuring huge quantity, diverse scenes and multiple motion types. HDRVD2K fills gaps of video training data and facilitate the development of LVC on HDR videos. Based on HDRVD2K, we further propose the first learned bit-depth scalable video compression (LBSVC) network for HDR videos by effectively exploiting bit-depth redundancy between videos of multiple dynamic ranges. To achieve this, we first propose a compression-friendly bit-depth enhancement module (BEM) to effectively predict original HDR videos based on compressed tone-mapped low dynamic range (LDR) videos and dynamic range prior, instead of reducing redundancy only through spatio-temporal predictions. Our method greatly improves the reconstruction quality and compression performance on HDR videos. Extensive experiments demonstrate the effectiveness of HDRVD2K on learned HDR video compression and great compression performance of our proposed LBSVC network. Code and dataset will be released in \href{https://github.com/sdkinda/HDR-Learned-Video-Coding}{https://github.com/sdkinda/HDR-Learned-Video-Coding}. 

\end{abstract}    
\section{Introduction}
\label{sec:intro}
\begin{figure}[!t]
	\centering
	\includegraphics[width=0.47\textwidth]{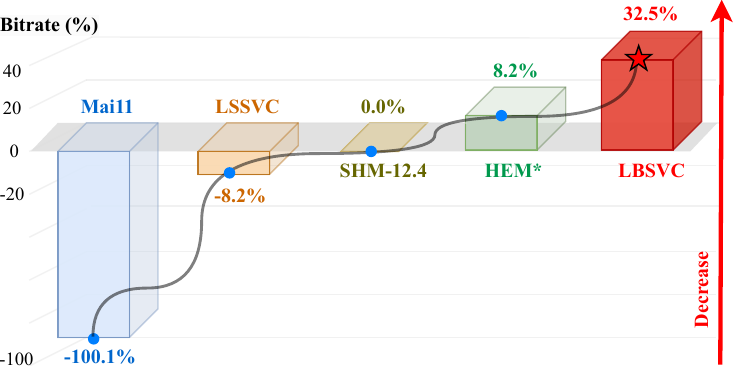}
	\caption{BD-Rate (PU-SSIM) comparisons with Mai11~\cite{mai2010optimizing}, LSSVC~\cite{10521480}, SHM~\cite{shvcmodel}, and HEM*~\cite{li2022hybrid}. The test dataset is HDM dataset~\cite{froehlich2014creating}. Our proposed LBSVC has a large performance drop under this setting.
	}
	\label{fig:fig1}
\end{figure}
In the past decade, high dynamic range (HDR) video capturing techniques and display devices have made remarkable progress and the demand for HDR video technology is increasing in various domains, such as photography, medical imaging, gaming. HDR videos can produce more realistic scenes on HDR displays with full visible light range compared with low dynamic range (LDR) videos ~\cite{Reinhard_book}. However, LDR videos typically employ a 24-bit (three 8-bit integer) per pixel encoding format, instead HDR videos represent natural scenes as floating point values. Common uncompressed floating point HDR content formats include .hdr~\cite{ward1994radiance}, .exr~\cite{bogart2003openexr} and (.tiff)~\cite{larson1998logluv}, in which .exr format~\cite{bogart2003openexr} even require total 48 bits for three color channels in one pixel. More bits contributes to HDR videos inevitably occupy huge transmission bandwidth and storage space, which hinders its widespread application.

In order to compress HDR video efficiently without compromising perceptual quality, several traditional HDR video compression algorithms have been proposed. 
They can be classified into two categories 1) single-layer perception-based algorithms and 2) two-layer scalable algorithms. \textbf{Single-layer perception-based algorithms}~\cite{mantiuk2004perception,5654069,5946532,7174544,zhang16mask,7805499,8239670,9628080} typically apply a perceptual quantization transformation to map HDR data to the maximum bit depth supported by the encoder and then compress the input according to the selected traditional codec. For compatibility with both LDR and HDR monitors, \textbf{two-layer scalable algorithms}~\cite{mantiuk2006backwards,mai2010optimizing, 6410792,koz2014methods,6525395,7786175,4378876,4378877,7572895,7475950,7532587}, also named backwards compatible algorithms, compress 8bit tone-mapped LDR video and then reconstruct 16bit floating point HDR video based on compressed LDR video.  

However, existing traditional HDR video compression methods rely heavily on the statistical properties of videos to optimize each hand-crafted modules, whose development has been limited. Therefore, it is necessary to explore further improvements with a whole compression system optimization. Recently, researchers~\cite{8953892,10204171,9941493,10655044} have been exploring learned video coding (LVC) based on deep neural networks to compress videos in a end-to-end way, which have achieved outstanding results and surpassed traditional codecs. Consequently, it is desirable to further improve HDR video compression performance with LVC methods by jointly optimizing the whole compression system.

To promote LVC methods on HDR videos, \textbf{the first issue} to be addressed urgently is the absence of training data. As far, there are some publicly accessible HDR video datasets~\cite{luthra2015call,froehlich2014creating,song2016sjtu,banitalebi2014quality}. These datasets are proposed to evaluate performance of HDR video compression or quality assessment algorithms, whose quantity, numbers of distinct scenes and motion patterns are limited. If LVC methods are trained on these methods, problems including insufficient model generalisation ability and lacking of diversity are inevitable. Hence, it is urgent for a large-scale HDR video dataset to promote LVC on HDR videos. Besides the datasets, \textbf{another key issue} is the inapplicability of existing LVC methods on HDR videos. Existing LVC methods~\cite{8953892,10204171,9941493,10655044} typically optimize their model on LDR videos and fail to consider the unique dynamic range characteristic of HDR videos, which achieve sub-optimal compression performance on HDR videos. Thus, novel LVC framework specifically designed for HDR videos are urgently needed.

Based on the above analysis, to facilitate the development of LVC on HDR videos, we build the first large-scale HDR video dataset, named HDRVD2K. To achieve high quality HDR video clips as training data, we collect 500 high-quality HDR videos and extract 2200 diverse video clips from these HDR videos. To verify the realism and dynamic range of these clips, careful subjective experiments on professional HDR display device are conducted. Meanwhile, we consider that selected clips should contain rich scenarios and different motion types, which are indispensable in video training datasets, so sufficient data analysis is conducted to verify the enrichment of HDRVD2K.

Based on HDRVD2K, we further propose the first learned bit-depth scalable video compression network (LBSVC) for HDR videos. In LBSVC, a dynamic range prior guided bit-depth enhancement module (BEM) is proposed to effectively predict HDR content based on compressed LDR videos and dynamic range prior. Specifically, BEM extracts dynamic range prior information from HDR videos  and compresses them losslessly with few bits. Then BEM utilizes dynamic range prior to enhance the dynamic range of compressed tone-mapped LDR video and predicts higher bit-depth HDR video. Hence, reconstruction quality of HDR video can be better and code stream to represent HDR video can be smaller. Experimental results show that our LBSVC can achieve 32.5\% bitrate saving over traditional scalable codec SHM~\cite{shvcmodel} in PU-SSIM metric~\cite{azimi2021pu21}, as shown in Fig.~\ref{fig:fig1}. In summary, our contributions are as follows:
\begin{itemize}
\item We are the first to propose a large-scale HDR video dataset named HDRVD2K, whose main features contain huge quantity, diverse scenes and multiple motion types. And proposed dataset fills the gap of training data on learned HDR video compression methods. 
\item We propose a LBSVC network for HDR videos, in which a BEM module is designed to effectively predict HDR video content based on compressed LDR video with the guidance of dynamic range prior, greatly improving the reconstruction quality of compressed HDR videos.
\item Extensive analysis and experiments demonstrate the superiority of our dataset and our method. Our work can become a new platform for researchers to explore LVC methods on HDR videos.
\end{itemize}

\section{Related Work}
\label{sec:relatedwork}


\subsection{Traditional HDR Video Compression}
\paragraph{Single-layer perception based method} 
Considering input data is typically	integer in digital image and video processing, high bit-depth floating-point data of HDR data should be transformed into low bit-depth integer data. Mantiuk et al.~\cite{mantiuk2004perception} first proposes a perception quantization scheme, which only requires only 10-11 bits to encode 12 orders of magnitude of visible luminance range, and then compresses HDR sequence with perception quantization transformation in MPEG-4. After that, a prominent perceptual quantizer (PQ) transfer function is presented in~\cite{miller2013perceptual}, which has the best perceptual uniformity in the considered luminance range~\cite{poynton2014perceptual} and has been selected as Anchor for HDR/WCG video coding~\cite{luthra2015call, 7174544}.  After that, some works to further promote single-layer perception based HDR video compression by optimizing PQ transfer function~\cite{zhang16mask,7805499,8239670, 9628080}. However, these methods fail to be compatible with LDR display device, limiting their spread to a certain extent.
\paragraph{Two-layer scalable method}
In order to be compatible with LDR technology and display devices, several backwards-compatible methods are proposed, which can compress both HDR content and its tone-mapped LDR version. They first utilize tone mapping operator (TMO) to map linear floating HDR content down to LDR bit-depths. Then tone-mapped LDR contents are compressed with a legacy encoder, referred to base layer (BL). Subsequently, these algorithms reconstruct HDR content (referred to enhancement layer, EL) with the information of BL. Mantiuk et al.~\cite{mantiuk2006backwards} propose a backward compatible HDR video compression method, which utilizes MPEG-4 to compress tone-mapped LDR version of HDR sequence and the residual between compressed tone-mapped LDR version and original HDR sequence. Subsequently, some researches focus on optimizing inter-layer prediction module to reduce the redundancy between compressed tone-mapped LDR and HDR video and improve reconstruction quality of HDR videos. However, these methods typically rely on the statistical properties of videos to optimize each hand-crafted modules and it is necessary to explore further improvements with a whole system optimization. 
\begin{figure*}[!t]
	\vspace{-0.4cm}
	\centering
	\includegraphics[width=\textwidth]{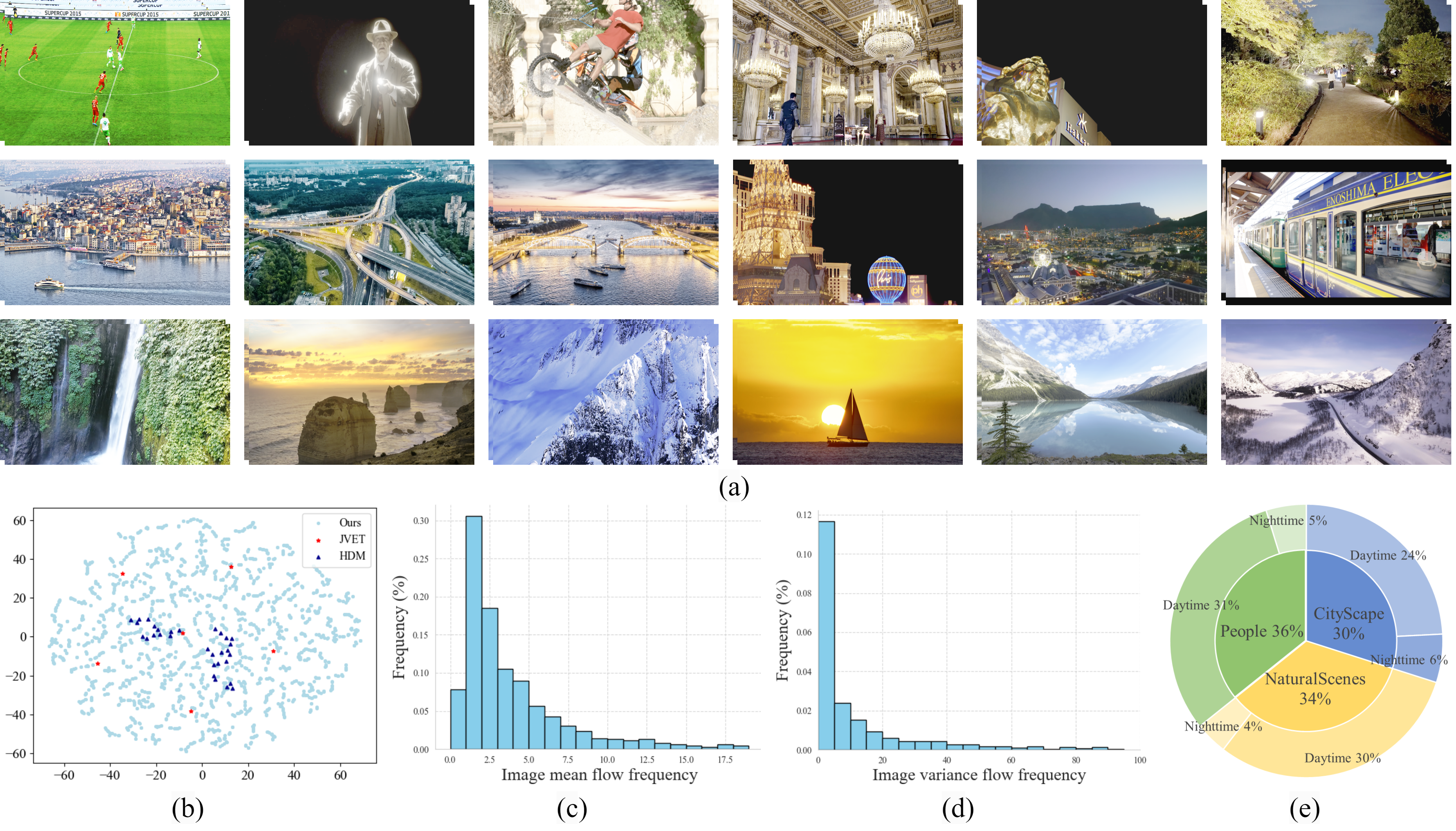}
	\vspace{-10pt}
	\caption{(a) Sampled video clips from HDRVD2K, which are all tome-mapped for visualization. (b) Diversity comparison among our dataset, JVET~\cite{luthra2015call} and HDM~\cite{froehlich2014creating}. (c) The histogram of flow mean magnitude of all pixels in the dataset. (d) The histogram of flow variance magnitude of all pixels in the dataset. (e) Scenes and illumination type percentage of HDRVD2K.
	}
	\label{dataset_samples}
	\vspace{-5pt}
\end{figure*}
\subsection{Learned Video Compression}
Recently, LVC has garnered significant interest from researchers, which optimize the whole compression sysytem in a end-to-end way. Existing single-layer video compression methods~\cite{8953892, 9880063, li2021deep, 9578150, 9941493, li2022hybrid, 10129217, 9667275, 10204171, 10655044} have seen rapid advancements in compression performance, even surpassing traditional video compression algorithms like H.266/VVC~\cite{9503377}. However, these methods focus on optimizing the compression efficiency of a single video signal, which is inadequate for meeting the demands of HDR video playback across different display devices. Scalable video coding schemes, on the other hand, offer greater flexibility and adaptability, supporting multiple resolutions~\cite{10521480} and tasks~\cite{9949576,10647798, 10222132} within their frameworks. This paper falls into this category and introduces the first learned bit-depth scalable scheme for HDR videos, effectively eliminating content redundancy between videos of varying bit depths (8-bit tone-mapped LDR and 16-bit HDR video).
\begin{table}[!t]
	\centering
	\caption{ Metrics to assess the diversity of different video datasets.}
	\vspace{-10pt}
	\renewcommand\arraystretch{1.2}
	\tabcolsep=0.1cm
	\begin{tabular}{c|c}
		\hline	
		\multicolumn{2}{c}{\textbf{Metrics on the extent of HDR}}\\ 
		\hline
		\textbf{FHLP} &  Fraction of HighLight Pixel: defined in~\cite{Guo_2023_CVPR} \\ \hline
		\textbf{EHL} & Extent of HighLight: defined in~\cite{Guo_2023_CVPR} \\ 
		\hline	
		\multicolumn{2}{c}{\textbf{ Metrics on the overall-style}}\\ 
		\hline
		\textbf{ALL} &  Average Luminance Level: defined in~\cite{Guo_2023_CVPR} \\ \hline
		\multirow{3}{*}{\textbf{DR}} & Dynamic Range~\cite{hu2022hdr}: calculated as the log10 \\
		&differences between the highest 2\% luminance \\ 
		& and the lowest 2\% luminance.  \\
		\hline
		\multicolumn{2}{c}{\textbf{ Metrics on intra-frame diversity}}\\ 
		\hline
		\textbf{SI} &   Spatial Information: defined in~\cite{siti} \\ 
		\hline
		\textbf{CF} &  Colorfulness: defined in~\cite{hasler2003measuring} \\ 
		\hline
		\textbf{stdL} &   standard deviation of Luminance: defined in~\cite{Guo_2023_CVPR} \\ 
		\hline
		\multicolumn{2}{c}{\textbf{ Metrics on the temporal motion}}\\ 
		\hline
		\textbf{TI} & Temporal Information: defined in~\cite{siti} \\ 
		\hline
	\end{tabular}
	\label{metric_dataset} 
	\vspace{-10pt}
\end{table}

\begin{table*}[!t]
	\centering
	\label{metrics} 
	\caption{ Statistics of different datasets.}
	\vspace{-10pt}
	\renewcommand\arraystretch{1.3}
	\tabcolsep=0.4cm
	\begin{tabular}{c|cc|ccc|cc|c}
		\hline
		\multirow{2}{*}{\diagbox{Dataset}{Metric}}
		&\multicolumn{2}{c|}{\textbf{Extent of HDR}}&\multicolumn{3}{c|}{\textbf{Intra-frame Diversity}}&\multicolumn{2}{c|}{\textbf{Overall-style}}&\textbf{Motion}\\
		\cline{2-9}
		&\textbf{FHLP$\uparrow$}&\textbf{EHL$\uparrow$}&\textbf{SI$\uparrow$}&\textbf{CF$\uparrow$}&\textbf{stdL$\uparrow$}&\textbf{ALL$\uparrow$}&\textbf{DR$\uparrow$}&\textbf{TI$\uparrow$}\\
		\hline
		JVET&1.10 &8.15 &52.09&16.96&6.24&2.73&\textbf{4.55}&11.24\\
		\hline
		HDM& 0.56& 4.69& 12.94& 5.14& 1.81& 0.97 & 2.82& 11.57 \\
		\hline
		HDRVD2K&\textbf{9.08}&\textbf{29.7}&\textbf{60.17}&\textbf{25.89}&\textbf{16.09}& \textbf{12.32}& 4.30&\textbf{14.58} \\
		\hline	
	\end{tabular}
\end{table*}
\subsection{HDR video dataset}
To evaluate various HDR image and video quality metrics, Banitalebi-Dehkordi et al.~\cite{banitalebi2014quality} propose a DML-HDR dataset. Froehlich et al.~\cite{froehlich2014creating} present a cinematic wide gamut HDR-video test set named HDM, designed for the evaluation of temporal tone mapping operators and HDR-displays. Luthra el al.~\cite{luthra2015call} propose a HDR video dataset as common test condition for HDR video compression experiments and assessment. Considering the majority of available datasets focusing HDR content are HD resolutions Song et al.~\cite{song2016sjtu} propose a UHD HDR video dataset named SJTU. However, these datasets are typically presented for testing and evaluation of HDR video compression methods, whose quantities and scenes are insufficient for LVC network training.
\section{Proposed Dataset}
\label{sec:dataset}
\paragraph{Construction of Our Dataset}
To favor the development of blooming LVC methods on HDR videos, we construct a large-scale HDR video dataset, named HDRVD2K. Actually, it is extremely challenging to collect large-scale and useful HDR video sequences, which is time-consuming and laborintensive. Former researchers have explained that the process of HDR video shooting can be very hard, resulting in HDR video datasets that are small in size and limited in terms of content. Alternatively, we resort to video platforms where many videos are taken with professional cameras.

Nowadays, there exists many high-quality HDR videos in website and we collect 500 videos in a professional HDR video format, which is SMPTE ST 2086, HDR10 compatible and whose color primaries are mainly BT.2020. In order to save storage space and accelerate the process of video editing, these videos are downloaded in a 3840$\times$2160 resolution with 60 fps, whose source resolution can be 4K or even 8K. After that, we utilize professional video editing software DaVinci Resolve Studio to edit downloaded HDR videos and generate useful HDR video clips. Specifically, during the 'EDIT' stage, we manually select clips whose duration are about 3 seconds and ensure the scenes and motions are distinctive in a video. Then we set the output gamma in color space transformation to Linear during the 'COLOR' stage in DaVinci Resolve Studio. After that, during the 'DELIVER' stage, output format of each clip is selected 'EXR'~\cite{bogart2003openexr} and the resolution is set to 1920$\times$1080, which aims to reduce the storage space of dataset.  Hence, thousands of video clips are generated and each clip contains 180 frames. 

To verify the dynamic range and realism of generated HDR video clips, some subjective tests are conducted. Following the ITU-R BT.500-13 Recommendation~\cite{series2012methodology}, the subjective tests have been conducted in a dark, quiet room, with the ambient illumination of the room at 2.154 lux and the luminance of the screen when turned off at 0.03 cd/m{$^2$}. The stimuli were presented on a calibrated HDR SIM2 HDR47ES4MB 47{$^\prime$$^\prime$} display~\cite{sim2} with 1920×1080 resolution, peak brightness of 4000 cd/m{$^2$}, used in its native HDR mode. The distance from the screen was fixed to three heights of the display, with the eyes of observers positioned zero degrees horizontally and vertically from the center of the display. Repeat and low-quality clips are discard during these tests. This way, the realism and dynamic range of generated video clips can be guaranteed.

As a result, we collect a new HDR video dataset, consisting of 500 videos with 2200 independent clips that are different from each other in content. To standard the input for training and testing, the resolution of all frames are fixed to 1920$\times$1080 and each clip randomly select 15 successive frames from generated 180 frames. Fig.~\ref{dataset_samples} (a) show some clips of our HDRVD2K, in which we can multiple contents including people, cityscapes and natural sceneries. For different scenes, we can observe scenes at different times including daytime and night time, whose distribution can be found in Fig.~\ref{dataset_samples} (e).
\begin{figure*}[!t]
		\vspace{-0.6cm}
	\centering
	\includegraphics[width=\textwidth]{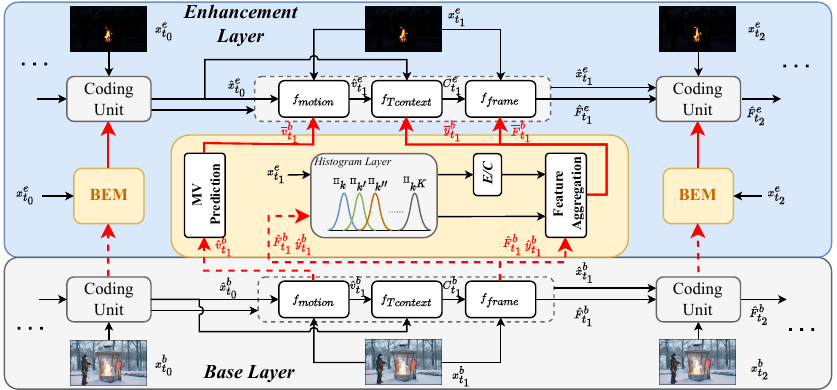}
	\caption{
		\textbf{Overview of our proposed LBSVC framework for HDR videos.} $f_{motion}$, $f_{Tcontext}$ and $f_{frame}$ in coding unit denote motion information compression, contextual information extraction and contextual information compression individually. $x_{t_i}^e$ and $x_{t_i}^b$indicate the i-th HDR frame and its tone-mapped LDR frame, which is the input of enhancement layer (EL) and base layer (BL). Code information in BL contains decoded LDR frame $\hat{x}_{t_i}^b$, decoded motion vector $\hat{v}_{t_i}^b$, decoded features ${F}_{t_i}^b$, decoded latent representation $\hat{y}_{t_i}^b$ and contextual information $C_{t_i}^b$. Code information in EL is similar with BL, with some external reference information from BEM module. E/C denotes dynamic range prior extraction and compression. 
	}
	\label{framework}
\end{figure*}
\paragraph{Analysis of Our Dataset} To quantitatively evaluate the superiority of our dataset, we analyze the diversity of the HDM dataset~\cite{froehlich2014creating}, JVET dataset~\cite{luthra2015call} and our dataset. Following~\cite{shu2024,Guo_2023_CVPR}, we utilize 8 metrics to assess the diversity of different HDR video datasets from the dimensions of intra-frame diversity, extent, overall style and temporal motion. For each video clip, 8 different metrics are calculated according to Table~\ref{metric_dataset}. Then we utilize the t-SNE~\cite{van2008visualizing} to project 8-D vector from Table~\ref{metric_dataset} of each video clip to the corresponding 2D-coordinate for plotting the dataset distribution of our dataset and comparison datasets. 

As shown in Fig.~\ref{dataset_samples} (b), our dataset contains wider frame distribution than JVET and HDM dataset, indicating that the networks trained with our dataset can be better generalized to different scenarios. Furthermore, we present the flow mean and variance magnitude in Fig.~\ref{metric_dataset} (c)(d), which demonstrate the motion diversity of our dataset. Detailed statistics of different datasets are shown in Table~\ref{metric_dataset} and we find our HDRVD2K outperforms other datasets on different assessment metrics. The diversity in both scenes and motion patterns makes that HDRVD2K can be used for training LVC methods and assessing the generalization capability of the networks across different scenes. More details about the dataset can be found in supplementary materials.
\section{Proposed Method}
\label{sec:method} 

\textbf{Overview} The coding structure of our proposed method is illustrated in Fig. \ref{framework}, whose input contains original 16-bit floating HDR video data {$X^e$}=\{$x_t^e|t=$1,...,n\} and its tone-mapped LDR data {$X^b$}=\{$x_t^b|t=$1,...,n\}. Firstly, LDR frames will be compressed in base layer (BL), which we use in LBSVC is a LVC method named DCVC-HEM~\cite{li2022hybrid}. And in enhancement layer (EL), original HDR frames will be compressed based on BL information and a bit-depth enhancement module (BEM). BEM utilizes dynamic range prior and code information from BL to predict HDR information effectively, which can improve the reconstruction quality of HDR video. More details of proposed LBSVC and specific framework can be found in supplementary materials.

\textbf{Base Layer Coding} BL coding pipeline contains three core steps: $f_{motion}$, $f_{Tcontext}$ and $f_{frame}$. $f_{motion}$ uses optical flow network to estimate the motion vector (MV) $v_t^b$, then $v_t^b$ is encoded and decoded as $\hat{v}_t^b$. Based on $\hat{v}_t^b$ and the propagated feature $\hat{F}_{t-1}^b$ from the previous frame, $f_{Tcontext}$ extracts the motion-aligned temporal context feature $C_t^b$. Finally, $f_{frame}$ encodes $x_{t}^b$ into quantized latent representation $\hat{y}_t^b$ and the output frame $\hat{x}_{t}^b$ is reconstructed via the decoder and frame generator after entropy coding. So far, tone-mapped LDR video is compressed with BL and then EL will compress original HDR video based on compressed LDR video.

\begin{figure}[!t]
	\centering
	\vspace{-0.6cm}
	\includegraphics[width=0.45\textwidth]{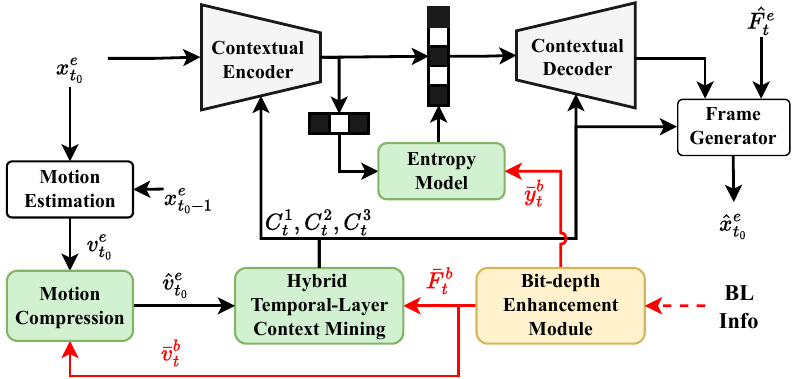}
	\caption{
		Architecture of EL coding unit.
	}
	\label{fig:ELCoding}
		\vspace{-0.5cm}
\end{figure}

\textbf{Enhancement Layer Coding} During the BL coding stage, tone-mapped LDR video frames are compressed and some code information are retained. Then EL compresses original HDR frame $x_{t_0}^e$ based on these code information and temporal previous compressed HDR frame information. Specifically, as shown in Fig.~\ref{fig:ELCoding} in EL, a motion estimation module is used to generate motion vectors information $v_{t_0}^e$ between $x_{t_0}^e$ and previous HDR frame $x_{t_0-1}^e$. Then a motion compression module compresses $v_{t_0}^e$ into bits based on $\overline{v}_{t}^b$ and decompress the bits into compressed motion vector information $\hat{v}_{t_0}^e$. After that, a hybrid temporal-layer context mining module in~\cite{10521480} is utilized to mine reconstructed temporal information ($\hat{v}_{t_0}^e$) and enhanced texture information ($\overline{F}_{t}^b$) as soon as possible to produce context hybrid $C_t^1,C_t^2,C_t^3$. The hybrid contexts $C_t^l$ (l=1,2,3) are then refilled into contextual encoder/decoder for the compression of HDR video frames in the EL. In contextual encoder/decoder, $\overline{y}_{t}^b$ will be feed into entropy model to estimate the latent code ${y}_{t}^e$ of HDR frame more effectively. Finally, reconstructed HDR frame $\hat{x}_{t_0}^e$ is generated with the frame generator module. Code information $\overline{v}_{t}^b$, $\overline{F}_{t}^b$ and $\overline{y}_{t}^b$ are the output of BEM module.

\begin{figure}[!t]
	\centering
		\vspace{-0.6cm}
	\includegraphics[width=0.45\textwidth]{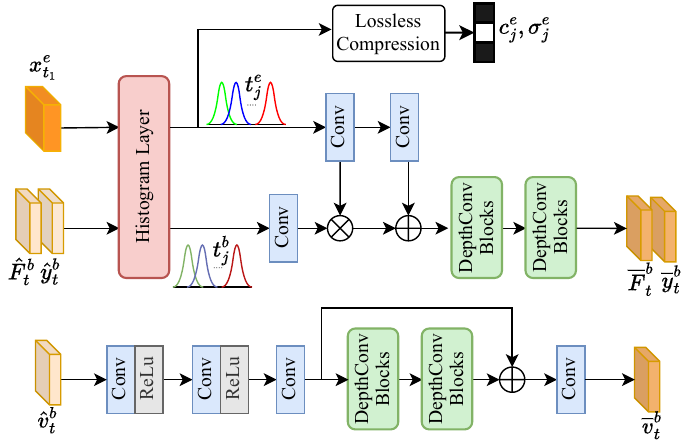}
	\vspace{-0.2cm}
	\caption{
		Architecture of the bit-depth enhancement module.
	}
	\label{fig:BEM}
		\vspace{-0.5cm}
\end{figure}

\textbf{Bit-depth Enhancement Module} HDR videos and corresponding tone-mapped LDR videos are similar in content. Hence, information of compressed tone-mapped LDR content in BL can predict information of HDR content in EL with few bits instead re-encoding HDR videos and inter-layer prediction is utilized. On the other hand, dynamic range of tone-mapped videos is vastly different from HDR videos. Hence, bit-depth enhancement technique is utilized to transform tone-mapped LDR videos to HDR videos. Traditional scalable compression methods use simple linear scaling~\cite{shvcmodel} or bit-consuming mapping~\cite{artusi2019overview} as bit-depth enhancement methods, which are inefficient to predict EL information with few bits. Instead, our BEM can predict EL information by combining BL information and dynamic range prior between HDR and tone-mapped LDR content. And dynamic range prior from HDR video can be represented with few bits. The detailed process of BEM is shown in Fig. \ref{fig:BEM}.

Specifically, BEM first extracts histogram information as dynamic range prior from HDR frame with a differential histogram layer in~\cite{avi2020deephist,hu2022hdr}, in which differential threshold functions $t_j(x)$ are utilized for the luminance range slicing and binning of input content.
\begin{equation}
	t_j(x) = exp\left(- \frac{\left(x-c_j\right)^2}{\sigma_j^2}\right) 
	\label{eq:1}
\end{equation}
where $c_j$ is the center of the sliced luminance range, and $\sigma_j$ represents the length of this range slice ($j = 1..k$). The BEM aggregates dynamic range prior $t_j^e$ extracted from HDR frames and threshold functions $t_j^b$ extracted from BL information ($\hat{F}_{t}^b,\hat{y}_t^b$), producing bit-depth enhanced output $\overline{F}_{t}^b,\overline{y}_t^b$. Dynamic range and content of output features are similar with corresponding features from original HDR content. Thus, fewer bits can be used to represent HDR contents with the assistance of enhanced features from BEM. Furthermore, dynamic range prior needs to be transmitted to decoder side, and we extract $c_j^e$ and $\sigma_j^e$ and compress them losslessly with additional few bits. The dynamic range prior compression of $c_j^e$ and $\sigma_j^e$ only costs 256 float-32 values while the prior can effectively represent the bin center and length of luminance range.

Code information $\hat{F}_{t}^b$ and $\hat{y}_t^b$ from BL have been enhanced with BEM for reducing the dynamic range difference between BL and EL. For another BL motion information $\hat{v}_{t}^b$, which represents compressed motion vector in BL, it can be similar with motion vector $\hat{v}_{t}^e$ in EL because motion in tone-mapped videos can be similar with original HDR videos. We utilize several depth convolution blocks, which have strong feature extraction performance and lightweight parameters, to minimize the representation redundancy between enhanced feature $\overline{v}_{t}^b$ and $\hat{v}_{t}^e$.
\begin{figure*}[!t]
	\vspace{-0.6cm}
	\centering
	\begin{subfigure}[b]{\textwidth}
		\includegraphics[width=\textwidth]{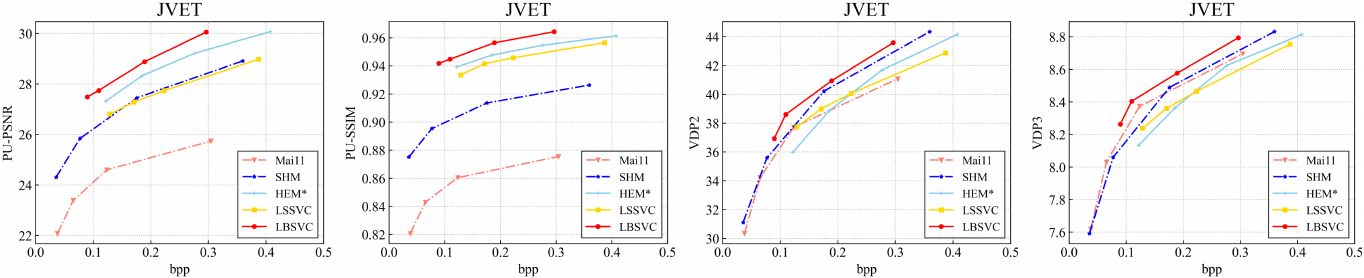} 
		\caption{BL+EL Rate-Distortion performance comparison on JVET dataset.}
	\end{subfigure}
	\begin{subfigure}[b]{\textwidth}
		\includegraphics[width=\textwidth]{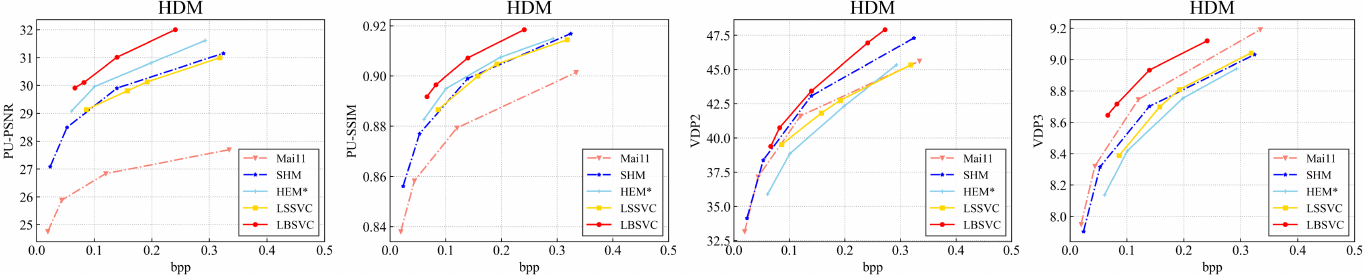} 
		\caption{BL+EL Rate-Distortion performance comparison on HDM dataset.}
	\end{subfigure}
	
	\caption{Rate and distortion curves for JVET and HDM datasets. Bpp is calculated by the sum of BL and EL. Four quality metrics are all calculated on HDR video frames. The intra period is set to 32. The metrics of HEM* tests on HDR video results and bpp is calculated based on two individual code streams.}
	\label{fig:exp_result}
\end{figure*}

\textbf{Loss function} In traditional bit-depth scalable video coding of SHM~\cite{shvcmodel}, a hierarchical
quality structure is applied to different layers, in which layers
are assigned different quantization parameters (QPs). For single BL and EL training stage in our method, loss function is similar with previous single-layer methods~\cite{9941493,li2022hybrid}:
\begin{equation}
		L_{single} = {\lambda}\cdot {D}\left(x_t^n,\hat{x}_t^n\right)+R_t^n 
			\label{eq:2}
\end{equation}
	
where ${D}\left(x_t^n,\hat{x}_t^n\right)$ denotes the distortion between original frame $x_t^n$ and reconstructed frame $\hat{x}_t^n$ in the n-th layer. The ${D}\left(\cdot\right)$ in this paper is the common mean squared error (MSE) distortion. The $R_t^n$ represents the bitrate estimated by the entropy model for encoding both MV and frame in the nth layer. For two-layer joint training stage, the loss function can be as follows:
\begin{align}
	\begin{split}
			L_{joint} = \left({\omega_b}\cdot{\lambda_b}\cdot {D}\left(x_t^b,\hat{x}_t^b\right)+R_t^b \right)\\
			+\left({\lambda_e}\cdot {D}\left(x_t^e,\hat{x}_t^e\right)+R_t^e \right),
	\end{split}
	\label{eq:joint}
\end{align}
where $\omega_b$ we set 0.5 in this paper.
\section{Experiments}
\label{exp}
\begin{table*}[!t] 
	\vspace{-0.2cm}
	\begin{center}   
		\normalsize
		\caption{BD-Rate (\%) on four metrics of different methods.}  
		\label{table:bdmetrics} 
		\vspace{-0.3cm}
		\begin{tabular}{ccccccccc}  
			\hline\hline   
			\multirow{2}{*}{Methods} & \multicolumn{2}{c}{PU-PSNR} & \multicolumn{2}{c}{PU-SSIM} & \multicolumn{2}{c}{VDP2}& \multicolumn{2}{c}{VDP3} \\   
			\cline{2-9}
			&JVET&HDM&JVET&HDM&JVET&HDM&JVET&HDM\\
			\hline
			SHM&0.0&0.0&0.0&0.0&0.0&0.0&0.0&0.0\\	
			Mai11&-&-&-&-&14.2&13.7&-6.6&-21.6\\
			HEM* & -30.6&-27.4   &-81.6&-9.4&29.1&66.6&28.0&39.2\\  
			LSSVC & 8.7&14.7&-69.2&8.2&30.4&43.4&31.3&13.4 \\  
			Ours & \textbf{-45.2}& \textbf{-49.8}&\textbf{-93.9}&\textbf{-32.5}&\textbf{-9.0}&\textbf{-7.9}&\textbf{-15.5}&\textbf{-44.2}\\ 
			\hline   
			\hline
		\end{tabular}   
	\end{center}   
	\vspace{-0.8cm}
\end{table*}

\textbf{Datasets} For training LVC methods on HDR videos, our HDRVD2K dataset is selected. Original 16-bit floating HDR video clips in dataset are used for EL stage and corresponding tone-mapped LDR version are used for BL stage. HDR and LDR video clips are all cropped into 256$\times$256 patches. During the test stage, to evaluate the compression performance of different methods on HDR videos, we select 6 sequences from JVET~\cite{luthra2015call} and 10 sequences from HDM~\cite{froehlich2014creating} as our test datasets, whose resolutions are all 1920$\times$1080. 

\textbf{Implementation Details} We first train the BL model~\cite{li2022hybrid} for different bitrates using $\lambda^b$= (85,170,380,840) to compress the tone-mapped LDR videos. After that, we freeze the parameters of BL and train EL model for four different bitrates using $\lambda^e$= (85,170,380,840). Finally, the loss function defined in Eq.~\ref{eq:joint} is utilized to finetune the training on two-layer model jointly. The Adam optimizer is used with default settings. Batch size is range from 1 to 16 with the change of training stage. The learning rate starts at ${1\times10}^{-4}$ and decays to ${1\times10}^{-5}$ in the later training stage. Both the BL and the EL models are implemented by Pytorch and trained on 1 Nvidia RTX4090. The training takes about 240 hours to finish.

\textbf{Settings} We follow~\cite{10521480} to encode 96 frames for each sequence in test datasets. Intra period is set to 32. Our comparison algorithms include latest reference software SHM-12.4 of SHVC\footnote{SHM-12.4: \url{https://vcgit.hhi.fraunhofer.de/jvet/SHM/-/tree/SHM-12.4}} ,a traditional backwards-compatible method Mai11~\cite{mai2010optimizing} combined with HEVC test model (HM-16.20)\footnote{HM16.20: \url{https://vcgit.hhi.fraunhofer.de/jvet/HM/-/tree/HM-16.20}}, and two LVC methods including a modified LSSVC network~\cite{10521480} for bit-depth scalable video coding and a method HEM* that encodes two layers separately based on ~\cite{li2022hybrid}. LSSVC is designed for spatial scalable coding and we modify some modules to fulfill the bit-depth scalable coding for HDR videos. In these four comparison algorithms, SHM, Mai11 and modified LSSVC are real bit-depth scalable methods, while HEM* method encode LDR and HDR videos separately for comparison, whose two code streams are unrelated. And follow existing LVC methods, our method and comparison algorithms are all tested on low-delay common test conditions. In SHM codec, we set the EL QP offsets $\Delta$QP = 4 because of its higher compression performance. Stable tone-mapping operator Mai11 is selected to transfer HDR videos to LDR videos for BL coding and 12bit-PQ coding is utilized on 16-bit floating HDR videos as preprocessing for EL coding.

\textbf{Evaluation Metrics} To evaluate the performance of two-layer HDR video coding, several quality assessment metrics are utilized. HDR-VDP-2~\cite{vdp2}, HDR-VDP-3~\cite{vdp3}, PU-PSNR and PU-SSIM are used for compressed HDR video results. PU-PSNR and PU-SSIM are computed after perceptually uniform encoding~\cite{azimi2021pu21}. When computing the HDR-VDP-2 and HDR-VDP-3, the diagonal display size is set to 21 inches and viewing distance in meters is set to 1. BD-Rate~\cite{bjontegaard2001calculation} calculates the bitrate difference between two methods under the same video quality and we utilize the same way to achieve four BD-Rates on above four metrics to evaluate overall compression quality of different methods. We abbreviate HDR-VDP-2 and HDR-VDP-3 as VDP2 and VDP3.

\textbf{Comparisons with SOTA Methods} 
Backward-compatible HDR video code streams can be decoded and then displayed on both LDR and HDR monitors, thus BL+EL performance is compared in our paper as the main comparison setting. Furthermore, BL performance on tone-mapped HDR videos utilize PSNR and BD-Rate as metrics. In BL+EL performance comparison, the bpp is calculated by the sum of BL and EL, and the distortion is calculated on HDR video frames. More details can be found in supplementary materials.

Fig.~\ref{fig:exp_result} presents the two-layer bit-depth scalable video coding performance on CTC and HDM datasets. It can be observed that our scheme outperforms other methods by a large margin on under all testing conditions, which effectively demonstrates the high compression efficiency of our method on HDR videos. For PU-PSNR and PU-SSIM metric, learned video compression methods achieve great performance because 12-bit PQ encoding transfers HDR content into perceptual uniformly domain~\cite{miller2013perceptual}, which helps leverage the strengths of LVC methods. LVC methods have achieved great compression performance on LDR videos and gamma corrected pixels of LDR videos are considered in perceptual uniformly domain~\cite{pu_metric}. On the other hand, we can find that PU-PSNR and PU-SSIM metric of Mai11~\cite{mai2010optimizing} are very low, because Mai11 utilizes piece-wise function mapping to reconstruct HDR videos based on tone-mapped videos. However, other methods optimize the EL coding on 12-bit PQ coding HDR videos with MSE loss in perceptual uniformly domain, so their PU metrics are higher than Mai11. 

LVC methods are no longer have the advantage on VDP2 and VDP3 metric because these metrics aim to evaluate images covering complete range of luminance the human eye can see, which are not the optimization direction of MSE Loss and PQ transfer function used in this paper. Our method utilize the high dynamic range prior in BEM to assist the reconstruction of HDR videos and thus we also achieve considerable performance on VDP2 and VDP3. To numerically visualize the overall rate-distortion performance of different methods, table~\ref{table:bdmetrics} is presented. It can be observed that our scheme surpasses all comparison algorithms and achieve the best compression performance under all test conditions, verifying the effectiveness of proposed method on HDR videos. 

Furthermore, in this paper, LVC methods~\cite{li2022hybrid, 10521480} are directly utilized to compress HDR videos with some modification and achieve considerable performance, which demonstrates the feasibility of LVC on HDR videos and further reflects the significance of proposed HDRVD2K. Our work bridges a platform for researchers to explore LVC methods on HDR videos.

\textbf{Complexity} 
Complexity of three scalable LVC methods are shown in Table \ref{complexity}. They are all calculated with BL+EL. HEM* are two individual HEM compression model so we double its complexity based on original HEM~\cite{li2022hybrid}. We can find the FLOPs and parameters of our LBSVC are similar to other LVC methods. However, our algorithm outperforms them greatly, which demonstrates the superiority of our method.
\begin{table}[!t] 
	\begin{center}   
		\normalsize
		\caption{Model complexity of different LVC methods.}  
		\label{complexity} 
		\vspace{-0.3cm}
		\begin{tabular}{ccc}  
			\hline\hline   Methods & Parameters (M) & Flops (G)  \\   
			\hline  
			HEM* & 35.04 &217.1 \\  
			LSSVC & 29.4 &246.48 \\  
			Ours & 41.8 &257.35\\ 
			\hline   
			\hline
		\end{tabular}   
	\end{center}   
	\vspace{-0.4cm}
\end{table}

\begin{table}[!t] 
	\vspace{-0.2cm}
	\begin{center}   
		\normalsize
		\setlength{\tabcolsep}{6pt}
		\caption{Ablation study of BEM with BD-rate (\%) on PU-PSNR and PU-SSIM.}  
		\label{table:ablation} 
		\vspace{-0.25cm}
			\begin{tabular}{ccccc}  
				\hline\hline   
				 \multirow{2}{*}{BEM} &	\multicolumn{2}{c}{PU-PSNR}& 	\multicolumn{2}{c}{PU-SSIM}\\
				 \cline{2-5}
				&JVET&HDM&JVET&HDM\\
				\hline
				\ding{51}&0.0&0.0&0.0&0.0\\
				\ding{55}&8.51&7.49&8.05&7.11\\
				\hline\hline
		\end{tabular}  
	\end{center}   
	\vspace{-0.9cm}
\end{table}

\textbf{Ablation Study} 
In our proposed LBSVC scheme, BEM is proposed to help reconstruct HDR videos with BL information and dynamic range prior. To verify
the effectiveness of the BEM, we conduct an ablation experiment to compare the
performance brought by BEM. As shown in Table \ref{table:ablation}, method with BEM is the anchor and for method without BEM, BD-Rate(PU-PSNR/PU-SSIM) will increase 8.51/7.49 and 8.05/7.11 on JVET and HDM dataset, which proves the effectiveness of BEM on reconstruction of HDR videos.
\section{Conclusion}
 In this paper, we propose a large scale dataset named HDRVD2K and a HDR video compression framework LBSVC. HDRVD2K features huge quantity, diverse scenes and multiple motion types and fills gaps of HDR video training data. Based on HDRVD2K, LBSVC is proposed to compress HDR videos, which effectively exploits bit-depth redundancy between videos of multiple dynamic ranges. To achieve this, a compression-friendly BEM is designed to effectively predict original HDR videos based on compressed tone-mapped LDR videos and dynamic range prior, greatly improving the reconstruction quality and compression performance on HDR videos. 
{
    \small
    \bibliographystyle{ieeenat_fullname}
    \bibliography{main}

\begin{thebibliography}{61}
\providecommand{\natexlab}[1]{#1}
\providecommand{\url}[1]{\texttt{#1}}
\expandafter\ifx\csname urlstyle\endcsname\relax
  \providecommand{\doi}[1]{doi: #1}\else
  \providecommand{\doi}{doi: \begingroup \urlstyle{rm}\Url}\fi

\bibitem[1788(2019)]{siti}
ITU-R Recommendation~BT. 1788.
\newblock Methodology for the subjective assessment of video quality in
  multimedia applications, 2019.

\bibitem[Artusi et~al.(2019)Artusi, Mantiuk, Richter, Hanhart, Korshunov,
  Agostinelli, Ten, and Ebrahimi]{artusi2019overview}
Alessandro Artusi, Rafa{\l}~K Mantiuk, Thomas Richter, Philippe Hanhart, Pavel
  Korshunov, Massimiliano Agostinelli, Arkady Ten, and Touradj Ebrahimi.
\newblock Overview and evaluation of the jpeg xt hdr image compression
  standard.
\newblock \emph{Journal of Real-Time Image Processing}, 16:\penalty0 413--428,
  2019.

\bibitem[Avi-Aharon et~al.(2020)Avi-Aharon, Arbelle, and
  Raviv]{avi2020deephist}
Mor Avi-Aharon, Assaf Arbelle, and Tammy~Riklin Raviv.
\newblock Deephist: Differentiable joint and color histogram layers for
  image-to-image translation.
\newblock \emph{arXiv preprint arXiv:2005.03995}, 2020.

\bibitem[Aydın et~al.(2008)Aydın, Mantiuk, and Seidel]{pu_metric}
Tun{\c{c}}~O. Aydın, Rafal Mantiuk, and Hans-Peter Seidel.
\newblock {Extending quality metrics to full luminance range images}.
\newblock In \emph{Human Vision and Electronic Imaging XIII}, page 68060B.
  International Society for Optics and Photonics, SPIE, 2008.

\bibitem[Azimi et~al.(2021)]{azimi2021pu21}
Maryam Azimi et~al.
\newblock Pu21: A novel perceptually uniform encoding for adapting existing
  quality metrics for hdr.
\newblock In \emph{2021 Picture Coding Symposium (PCS)}, pages 1--5. IEEE,
  2021.

\bibitem[Banitalebi-Dehkordi et~al.(2014)Banitalebi-Dehkordi, Azimi, Dong,
  Pourazad, and Nasiopoulos]{banitalebi2014quality}
A Banitalebi-Dehkordi, M Azimi, Y Dong, MT Pourazad, and P Nasiopoulos.
\newblock Quality assessment of high dynamic range (hdr) video content using
  existing full-reference metrics.
\newblock \emph{ISO/IEC JTC1/SC29/WG11, France}, 2014.

\bibitem[Bian et~al.(2024)Bian, Sheng, Li, and Liu]{10521480}
Yifan Bian, Xihua Sheng, Li Li, and Dong Liu.
\newblock Lssvc: A learned spatially scalable video coding scheme.
\newblock \emph{IEEE Trans. Image Process.}, 33:\penalty0 3314--3327, 2024.

\bibitem[Bj{\o}ntega(2001)]{bjontegaard2001calculation}
G Bj{\o}ntega.
\newblock Calculation of average psnr differences between rdcurves.
\newblock document VCEG-M33, 2001.

\bibitem[Bogart et~al.(2003)Bogart, Kainz, and Hess]{bogart2003openexr}
R Bogart, F Kainz, and D Hess.
\newblock Openexr image file format.
\newblock In \emph{ACM SIGGRAPH}, page~28, 2003.

\bibitem[Bross et~al.(2021)Bross, Wang, Ye, Liu, Chen, Sullivan, and
  Ohm]{9503377}
Benjamin Bross, Ye-Kui Wang, Yan Ye, Shan Liu, Jianle Chen, Gary~J. Sullivan,
  and Jens-Rainer Ohm.
\newblock Overview of the versatile video coding (vvc) standard and its
  applications.
\newblock \emph{IEEE Trans. Circuit Syst. Video Technol.}, 31\penalty0
  (10):\penalty0 3736--3764, 2021.

\bibitem[Chen et~al.(2015)Chen, Boyce, Ye, and Hannuksela]{shvcmodel}
J Chen, J Boyce, Y Ye, and MM Hannuksela.
\newblock Scalable hevc (shvc) test model 10 (shm 10).
\newblock \emph{JCT-VC of ITU-T SG16 WP 3and ISO/IEX JTC}, 1, 2015.

\bibitem[Chen et~al.(2024)Chen, Gao, and Liu]{10647798}
Qiaoxi Chen, Changsheng Gao, and Dong Liu.
\newblock End-to-end learned scalable multilayer feature compression for
  machine vision tasks.
\newblock In \emph{IEEE Int. Conf. Image Process. (ICIP)}, pages 1781--1787,
  2024.

\bibitem[Choi and Bajić(2022)]{9949576}
Hyomin Choi and Ivan~V. Bajić.
\newblock Scalable video coding for humans and machines.
\newblock In \emph{2022 IEEE 24th International Workshop on Multimedia Signal
  Processing (MMSP)}, pages 1--6, 2022.

\bibitem[Corporation(2016)]{sim2}
SIM2 Corporation.
\newblock Sim2 hdr47 monitors, 2016.
\newblock \href{http://www.sim2.com/}{http://www.sim2.com/}.

\bibitem[François et~al.(2016)François, Fogg, He, Li, Luthra, and
  Segall]{7174544}
Edouard François, Chad Fogg, Yuwen He, Xiang Li, Ajay Luthra, and Andrew
  Segall.
\newblock High dynamic range and wide color gamut video coding in hevc: Status
  and potential future enhancements.
\newblock \emph{IEEE Trans. Circuit Syst. Video Technol.}, 26\penalty0
  (1):\penalty0 63--75, 2016.

\bibitem[Froehlich et~al.(2014)Froehlich, Grandinetti, Eberhardt, Walter,
  Schilling, and Brendel]{froehlich2014creating}
Jan Froehlich, Stefan Grandinetti, Bernd Eberhardt, Simon Walter, Andreas
  Schilling, and Harald Brendel.
\newblock Creating cinematic wide gamut hdr-video for the evaluation of tone
  mapping operators and hdr-displays.
\newblock In \emph{Digital photography X}, pages 279--288. SPIE, 2014.

\bibitem[Garbas and Thoma(2011)]{5946532}
Jens-Uwe Garbas and Herbert Thoma.
\newblock Temporally coherent luminance-to-luma mapping for high dynamic range
  video coding with h.264/avc.
\newblock In \emph{2011 IEEE International Conference on Acoustics, Speech and
  Signal Processing (ICASSP)}, pages 829--832, 2011.

\bibitem[Guo et~al.(2023)Guo, Fan, Xue, and Jiang]{Guo_2023_CVPR}
Cheng Guo, Leidong Fan, Ziyu Xue, and Xiuhua Jiang.
\newblock Learning a practical sdr-to-hdrtv up-conversion using new dataset and
  degradation models.
\newblock In \emph{IEEE Conf. Comput. Vis. Pattern Recog. (CVPR)}, pages
  22231--22241, 2023.

\bibitem[Hasler and Suesstrunk(2003)]{hasler2003measuring}
David Hasler and Sabine~E Suesstrunk.
\newblock Measuring colorfulness in natural images.
\newblock In \emph{Human vision and electronic imaging VIII}, pages 87--95.
  SPIE, 2003.

\bibitem[Hu et~al.(2022{\natexlab{a}})Hu, Shen, Jiang, Ma, and An]{hu2022hdr}
Xiangyu Hu, Liquan Shen, Mingxing Jiang, Ran Ma, and Ping An.
\newblock La-hdr: Light adaptive hdr reconstruction framework for single ldr
  image considering varied light conditions.
\newblock \emph{IEEE Trans. Multimedia}, 25:\penalty0 4814--4829,
  2022{\natexlab{a}}.

\bibitem[Hu et~al.(2021)Hu, Lu, and Xu]{9578150}
Zhihao Hu, Guo Lu, and Dong Xu.
\newblock Fvc: A new framework towards deep video compression in feature space.
\newblock In \emph{IEEE Conf. Comput. Vis. Pattern Recog. (CVPR)}, pages
  1502--1511, 2021.

\bibitem[Hu et~al.(2022{\natexlab{b}})Hu, Lu, Guo, Liu, Jiang, and Xu]{9880063}
Zhihao Hu, Guo Lu, Jinyang Guo, Shan Liu, Wei Jiang, and Dong Xu.
\newblock Coarse-to-fine deep video coding with hyperprior-guided mode
  prediction.
\newblock In \emph{IEEE Conf. Comput. Vis. Pattern Recog. (CVPR)}, pages
  5911--5920, 2022{\natexlab{b}}.

\bibitem[Jiang et~al.(2023)Jiang, Choi, Racapé, Feltman, and
  Kamisli]{10222132}
Wei Jiang, Hyomin Choi, Fabien Racapé, Simon Feltman, and Fatih Kamisli.
\newblock Face restoration-based scalable quality coding for video
  conferencing.
\newblock In \emph{2023 IEEE International Conference on Multimedia and Expo
  Workshops (ICMEW)}, pages 206--211, 2023.

\bibitem[Jin et~al.(2023)Jin, Lei, Peng, Pan, Li, and Ling]{10129217}
Dengchao Jin, Jianjun Lei, Bo Peng, Zhaoqing Pan, Li Li, and Nam Ling.
\newblock Learned video compression with efficient temporal context learning.
\newblock \emph{IEEE Trans. Image Process.}, 32:\penalty0 3188--3198, 2023.

\bibitem[Koz and Dufau(2012)]{6410792}
Alper Koz and Frederic Dufau.
\newblock Optimized tone mapping with perceptually uniform luminance values for
  backward-compatible high dynamic range video compression.
\newblock In \emph{IEEE Int. Conf. Vis. Commun. Image Process. (VCIP)}, pages
  1--6, 2012.

\bibitem[Koz and Dufaux(2014)]{koz2014methods}
Alper Koz and Frederic Dufaux.
\newblock Methods for improving the tone mapping for backward compatible high
  dynamic range image and video coding.
\newblock \emph{Signal Processing: Image Communication}, 29\penalty0
  (2):\penalty0 274--292, 2014.

\bibitem[Larson(1998)]{larson1998logluv}
Gregory~Ward Larson.
\newblock Logluv encoding for full-gamut, high-dynamic range images.
\newblock \emph{Journal of Graphics Tools}, 3\penalty0 (1):\penalty0 15--31,
  1998.

\bibitem[Lasserre et~al.(2016)Lasserre, Léannec, Poirier, and Galpin]{7786175}
Sébastien Lasserre, Fabrice~Le Léannec, Tangi Poirier, and Franck Galpin.
\newblock Backward compatible hdr video compression system.
\newblock In \emph{2016 Data Compression Conference (DCC)}, pages 309--318,
  2016.

\bibitem[Le~Pendu et~al.(2016)Le~Pendu, Guillemot, and Thoreau]{7475950}
Mikaël Le~Pendu, Christine Guillemot, and Dominique Thoreau.
\newblock Inter-layer prediction of color in high dynamic range image scalable
  compression.
\newblock \emph{IEEE Trans. Image Process.}, 25\penalty0 (8):\penalty0
  3585--3596, 2016.

\bibitem[Li et~al.(2021)Li, Li, and Lu]{li2021deep}
Jiahao Li, Bin Li, and Yan Lu.
\newblock Deep contextual video compression.
\newblock \emph{Adv. Neural Inform. Process. Syst. (NIPS)}, 34:\penalty0
  18114--18125, 2021.

\bibitem[Li et~al.(2022)Li, Li, and Lu]{li2022hybrid}
Jiahao Li, Bin Li, and Yan Lu.
\newblock Hybrid spatial-temporal entropy modelling for neural video
  compression.
\newblock In \emph{ACM Int. Conf. Multimedia (ACMMM)}, pages 1503--1511, 2022.

\bibitem[Li et~al.(2023)Li, Li, and Lu]{10204171}
Jiahao Li, Bin Li, and Yan Lu.
\newblock Neural video compression with diverse contexts.
\newblock In \emph{IEEE Conf. Comput. Vis. Pattern Recog. (CVPR)}, pages
  22616--22626, 2023.

\bibitem[Li et~al.(2024)Li, Li, and Lu]{10655044}
Jiahao Li, Bin Li, and Yan Lu.
\newblock Neural video compression with feature modulation.
\newblock In \emph{IEEE Conf. Comput. Vis. Pattern Recog. (CVPR)}, pages
  26099--26108, 2024.

\bibitem[Liu et~al.(2019)Liu, Sidaty, Hamidouche, Déforges, Valenzise, and
  Zerman]{8239670}
Yi Liu, Naty Sidaty, Wassim Hamidouche, Olivier Déforges, Giuseppe Valenzise,
  and Emin Zerman.
\newblock An adaptive quantizer for high dynamic range content: Application to
  video coding.
\newblock \emph{IEEE Trans. Circuit Syst. Video Technol.}, 29\penalty0
  (2):\penalty0 531--545, 2019.

\bibitem[Liu et~al.(2022)Liu, Sidaty, Hamidouche, Déforges, and Jung]{9628080}
Yi Liu, Naty Sidaty, Wassim Hamidouche, Olivier Déforges, and Cheolkon Jung.
\newblock Visual attention-aware high dynamic range quantization for hevc video
  coding.
\newblock \emph{IEEE Trans. Circuit Syst. Video Technol.}, 32\penalty0
  (7):\penalty0 4296--4311, 2022.

\bibitem[Lu et~al.(2019)Lu, Ouyang, Xu, Zhang, Cai, and Gao]{8953892}
Guo Lu, Wanli Ouyang, Dong Xu, Xiaoyun Zhang, Chunlei Cai, and Zhiyong Gao.
\newblock Dvc: An end-to-end deep video compression framework.
\newblock In \emph{IEEE Conf. Comput. Vis. Pattern Recog. (CVPR)}, pages
  10998--11007, 2019.

\bibitem[Luthra et~al.(2015)Luthra, Francois, and Husak]{luthra2015call}
A Luthra, E Francois, and W Husak.
\newblock Call for evidence (cfe) for hdr and wcg video coding, document
  n15083, iso.
\newblock Technical report, ISO/IEC JTC1/SC29/WG11, Geneva, Switzerland, 2015.

\bibitem[Mai et~al.(2010)Mai, Mansour, Mantiuk, Nasiopoulos, Ward, and
  Heidrich]{mai2010optimizing}
Zicong Mai, Hassan Mansour, Rafal Mantiuk, Panos Nasiopoulos, Rabab Ward, and
  Wolfgang Heidrich.
\newblock Optimizing a tone curve for backward-compatible high dynamic range
  image and video compression.
\newblock \emph{IEEE Trans. Image Process.}, 20\penalty0 (6):\penalty0
  1558--1571, 2010.

\bibitem[Mai et~al.(2013)Mai, Mansour, Nasiopoulos, and Ward]{6525395}
Zicong Mai, Hassan Mansour, Panos Nasiopoulos, and Rabab~Kreidieh Ward.
\newblock Visually favorable tone-mapping with high compression performance in
  bit-depth scalable video coding.
\newblock \emph{IEEE Trans. Multimedia}, 15\penalty0 (7):\penalty0 1503--1518,
  2013.

\bibitem[Mantiuk et~al.(2004)Mantiuk, Krawczyk, Myszkowski, and
  Seidel]{mantiuk2004perception}
Rafal Mantiuk, Grzegorz Krawczyk, Karol Myszkowski, and Hans-Peter Seidel.
\newblock Perception-motivated high dynamic range video encoding.
\newblock \emph{ACM Trans. Graph.}, 23\penalty0 (3):\penalty0 733--741, 2004.

\bibitem[Mantiuk et~al.(2006)Mantiuk, Efremov, Myszkowski, and
  Seidel]{mantiuk2006backwards}
Rafa\l{} Mantiuk, Alexander Efremov, Karol Myszkowski, and Hans-Peter Seidel.
\newblock Backward compatible high dynamic range mpeg video compression.
\newblock In \emph{ACM SIGGRAPH}, page 713–723, New York, NY, USA, 2006.
  Association for Computing Machinery.

\bibitem[Mantiuk et~al.(2011)Mantiuk, Kim, Rempel, and Heidrich]{vdp2}
Rafa\l{} Mantiuk, Kil~Joong Kim, Allan~G. Rempel, and Wolfgang Heidrich.
\newblock Hdr-vdp-2: a calibrated visual metric for visibility and quality
  predictions in all luminance conditions.
\newblock \emph{ACM Trans. Graph.}, 30\penalty0 (4), 2011.

\bibitem[Mantiuk et~al.(2023)Mantiuk, Hammou, and Hanji]{vdp3}
Rafal~K Mantiuk, Dounia Hammou, and Param Hanji.
\newblock Hdr-vdp-3: A multi-metric for predicting image differences, quality
  and contrast distortions in high dynamic range and regular content.
\newblock \emph{arXiv preprint arXiv:2304.13625}, 2023.

\bibitem[Miller et~al.(2013)Miller, Nezamabadi, and Daly]{miller2013perceptual}
Scott Miller, Mahdi Nezamabadi, and Scott Daly.
\newblock Perceptual signal coding for more efficient usage of bit codes.
\newblock \emph{SMPTE Motion Imaging Journal}, 122\penalty0 (4):\penalty0
  52--59, 2013.

\bibitem[Mir et~al.(2016)Mir, Talagala, Arachchi, and Fernando]{7532587}
Junaid Mir, Dumidu~S. Talagala, Hemantha~Kodikara Arachchi, and Anil Fernando.
\newblock Adaptive residual mapping for an efficient extension layer coding in
  two-layer hdr video coding.
\newblock In \emph{IEEE Int. Conf. Image Process. (ICIP)}, pages 1394--1398,
  2016.

\bibitem[Motra and Thoma(2010)]{5654069}
Ajit Motra and Herbert Thoma.
\newblock An adaptive logluv transform for high dynamic range video
  compression.
\newblock In \emph{IEEE Int. Conf. Image Process. (ICIP)}, pages 2061--2064,
  2010.

\bibitem[Poynton and Funt(2014)]{poynton2014perceptual}
Charles Poynton and Brian Funt.
\newblock Perceptual uniformity in digital image representation and display.
\newblock \emph{Color Research \& Application}, 39\penalty0 (1):\penalty0
  6--15, 2014.

\bibitem[Reinhard et~al.(2005)Reinhard, Ward, Pattanaik, and
  Debevec]{Reinhard_book}
Erik Reinhard, Greg Ward, Sumanta Pattanaik, and Paul Debevec.
\newblock \emph{High Dynamic Range Imaging: Acquisition, Display, and
  Image-Based Lighting (The Morgan Kaufmann Series in Computer Graphics)}.
\newblock Morgan Kaufmann Publishers Inc., San Francisco, CA, USA, 2005.

\bibitem[Reinhard et~al.(2010)Reinhard, Ward, Pattanaik, and
  Debevec]{shvcmodel2}
E Reinhard, G Ward, S Pattanaik, and P Debevec.
\newblock Scalable hevc (shvc) test model 9 (shm 9), 2010.

\bibitem[Segall(2007)]{4378876}
Andrew Segall.
\newblock Scalable coding of high dynamic range video.
\newblock In \emph{IEEE Int. Conf. Image Process. (ICIP)}, pages I -- 1--I --
  4, 2007.

\bibitem[Series(2012)]{series2012methodology}
B Series.
\newblock Methodology for the subjective assessment of the quality of
  television pictures.
\newblock \emph{Recommendation ITU-R BT}, 500\penalty0 (13), 2012.

\bibitem[Sheng et~al.(2023)Sheng, Li, Li, Li, Liu, and Lu]{9941493}
Xihua Sheng, Jiahao Li, Bin Li, Li Li, Dong Liu, and Yan Lu.
\newblock Temporal context mining for learned video compression.
\newblock \emph{IEEE Trans. Multimedia}, 25:\penalty0 7311--7322, 2023.

\bibitem[Shu et~al.(2024)Shu, Shen, Hu, Li, and Zhou]{shu2024}
Yong Shu, Liquan Shen, Xiangyu Hu, Mengyao Li, and Zihao Zhou.
\newblock Towards real-world hdr video reconstruction: A large-scale benchmark
  dataset and a two-stage alignment network.
\newblock In \emph{IEEE Conf. Comput. Vis. Pattern Recog. (CVPR)}, pages
  2879--2888, 2024.

\bibitem[Song et~al.(2016)Song, Liu, Yang, Zhai, Xie, and Zhang]{song2016sjtu}
Li Song, Yankai Liu, Xiaokang Yang, Guangtao Zhai, Rong Xie, and Wenjun Zhang.
\newblock The sjtu hdr video sequence dataset.
\newblock In \emph{Proceedings of International Conference on Quality of
  Multimedia Experience (QoMEX 2016)}, page 100, 2016.

\bibitem[Van~der Maaten and Hinton(2008)]{van2008visualizing}
Laurens Van~der Maaten and Geoffrey Hinton.
\newblock Visualizing data using t-sne.
\newblock \emph{Journal of machine learning research}, 9\penalty0 (11), 2008.

\bibitem[Ward(1994)]{ward1994radiance}
Gregory~J Ward.
\newblock The radiance lighting simulation and rendering system.
\newblock In \emph{Proc. 21st Annu. Conf. Comput. Graph. Interact. Techn.},
  pages 459--472, 1994.

\bibitem[Wei et~al.(2018)Wei, Wen, and Li]{7572895}
Zhe Wei, Changyun Wen, and Zhengguo Li.
\newblock Local inverse tone mapping for scalable high dynamic range image
  coding.
\newblock \emph{IEEE Trans. Circuit Syst. Video Technol.}, 28\penalty0
  (2):\penalty0 550--555, 2018.

\bibitem[Winken et~al.(2007)Winken, Marpe, Schwarz, and Wiegand]{4378877}
Martin Winken, Detlev Marpe, Heiko Schwarz, and Thomas Wiegand.
\newblock Bit-depth scalable video coding.
\newblock In \emph{IEEE Int. Conf. Image Process. (ICIP)}, pages I -- 5--I --
  8, 2007.

\bibitem[Yu et~al.(2016)Yu, Jung, and Ke]{7805499}
Shengtao Yu, Cheolkon Jung, and Peng Ke.
\newblock Adaptive pq: Adaptive perceptual quantizer for hevc main 10
  profile-based hdr video coding.
\newblock In \emph{IEEE Int. Conf. Vis. Commun. Image Process. (VCIP)}, pages
  1--4, 2016.

\bibitem[Yılmaz and Tekalp(2022)]{9667275}
M.~Akın Yılmaz and A.~Murat Tekalp.
\newblock End-to-end rate-distortion optimized learned hierarchical
  bi-directional video compression.
\newblock \emph{IEEE Trans. Image Process.}, 31:\penalty0 974--983, 2022.

\bibitem[Zhang et~al.(2016)Zhang, Naccari, Agrafiotis, Mrak, and
  Bull]{zhang16mask}
Yang Zhang, Matteo Naccari, Dimitris Agrafiotis, Marta Mrak, and David~R. Bull.
\newblock High dynamic range video compression exploiting luminance masking.
\newblock \emph{IEEE Trans. Circuit Syst. Video Technol.}, 26\penalty0
  (5):\penalty0 950--964, 2016.

\end{thebibliography}
}


\end{document}